# Analysis and visualisation of RDF resources in Ondex

Catherine Canevet[1], Artem Lysenko[1], Andrea Splendiani[1*], Matthew Pocock[2], Christopher Rawlings[1]

[1] Centre for Mathematical and Computational Biology, Rothamsted Research, Harpenden, UK
[2] School of Computing Science, University of Newcastle, Newcastle, UK

An increasing number of biomedical resources provide their information on the Semantic Web and this creates the basis for a distributed knowledge base which has the potential to advance biomedical research [1]. This potential, however, cannot be realised until researchers from the life sciences can interact with information in the Semantic Web. In particular, there is a need for tools that provide data reduction, visualization and interactive analysis capabilities.

Ondex is a data integration and visualization platform developed to support Systems Biology Research [2]. At its core is a data model based on two main principles: first, all information can be represented as a graph and, second, all elements of the graph can be annotated with ontologies. This data model is conformant to the Semantic Web framework, in particular to RDF, and therefore Ondex is ideally positioned as a platform that can exploit the semantic web.

The Ondex system offers a range of features and analysis methods of potential value to semantic web users, including:
- An interactive graph visualization interface (Ondex user client), which provides data reduction and representation methods that leverage the ontological annotation.
- A suite of importers from a variety of data sources to Ondex (http://ondex.org/formats.html)
- A collection of plug-ins which implement graph analysis, graph transformation and graph-matching functions.
- An integration toolkit (Ondex Integrator) which allows users to compose workflows from these modular components
- In addition, all importers and plug-ins are available as web-services which can be integrated in other tools, as for instance Taverna [3].

The developments that will be presented in this demo have made this functionality interoperable with the Semantic Web framework. In particular we have developed an interactive importer, based on SPARQL that allows the query-driven construction of datasets which brings together information from different RDF data resources into Ondex (Fig. 1).

These datasets can then be further refined, analysed and annotated both interactively using the Ondex user client and via user-defined workflows. The results of these analyses can be exported in RDF, which can be used to enrich existent knowledge bases, or to provide application-specific views of the data. Both importer and exporter only focus on a subset of the Ondex and RDF data models, which are shared between these two data representations [4].

In this demo we will show how Ondex can be used to query, analyse and visualize Semantic Web knowledge bases. In particular we will present real use cases focused, but not limited to, resources relevant to plant biology.

We believe that Ondex can be a valid contribution to the adoption of the Semantic Web in Systems Biology research and in biomedical investigation more generally. We welcome feedback on our current import/export prototype and suggestions for the advancement of Ondex for the Semantic Web.

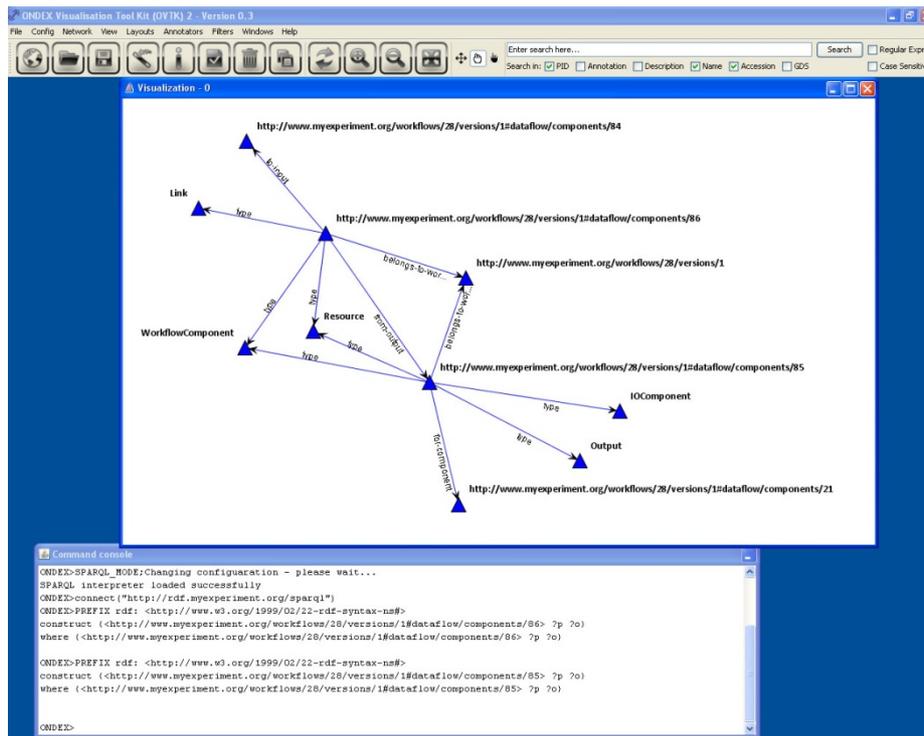

Figure 1: A screenshot from the Ondex RDF importer, showing an interactive console through which users can specify SPARQL queries. Results of these SPARQL queries is directly imported into Ondex and visualized.

## References

1. Ruttenberg, A. et. al.: Advancing translational research with the Semantic Web, BMC Bioinformatics, 8 (Suppl. 3): S2 (2007).
2. Köhler, J., Baumbach, J., Taubert, J., Specht, M., Skusa, A., Ruegg, A., Rawlings, C., Verrier, P., Philippi, S.: Graph-based analysis and visualization of experimental results with Ondex. Bioinformatics 22 (11):1383-1390 (2006).
3. Rawlings, C.: Semantic Data Integration for Systems Biology Research, Technology Track at ISMB'09, http://www.iscb.org/uploaded/css/36/11846.pdf (2009).
4. Splendiani, A. et. al.: Ondex semantic definition, (Web document) http://ondex.svn.sourceforge.net/viewvc/ondex/trunk/doc/semantics/ (2009).